\PassOptionsToPackage{usenames,dvipsnames,table}{xcolor}
\documentclass[11pt,a4paper]{article}
\usepackage[hyperref]{emnlp2020}
\usepackage{times}
\usepackage{latexsym}

\usepackage{microtype}

\usepackage{textcomp}
\usepackage{url}

\aclfinalcopy % Uncomment this line for the final submission
\def\aclpaperid{3010} %  Enter the acl Paper ID here

\usepackage{times}
\usepackage{latexsym}
\usepackage{amssymb}
\usepackage{array}
\usepackage{algorithm}
\usepackage{algpseudocode}
\usepackage{graphicx}
\usepackage{url}
\usepackage{times}
\usepackage{helvet}
\usepackage{courier}
\usepackage{balance}  % for  \balance command ON LAST PAGE  (only there!)
\usepackage{mdwlist}
\usepackage{graphics}
\usepackage{color}
\usepackage{rotating}
\usepackage{booktabs}
\usepackage{alltt}
\usepackage{tabularx}%http://ctan.org/pkg/tabularx
\usepackage{cellspace}
\usepackage{moreverb}
\usepackage{fancyvrb}
\usepackage{enumerate}
\usepackage{xspace}
\usepackage{relsize}
\usepackage{amsfonts}
\usepackage{savesym}
\usepackage{amsmath}
\savesymbol{iint}
\usepackage{txfonts}
\restoresymbol{TXF}{iint}

\usepackage[labelfont=bf]{caption}
\usepackage{dsfont}
\usepackage{rotating}
\usepackage{latexsym}
\usepackage{float}
\usepackage{enumitem}
\usepackage{subfig}
\usepackage{arydshln}
\usepackage{setspace}
\usepackage[page]{appendix}
\usepackage{seqsplit}
\usepackage{mathptmx}
\usepackage{anyfontsize}
\usepackage{t1enc}
\usepackage{adjustbox}
\usepackage{chngpage}
\usepackage{bbm}
\usepackage{lipsum}

\usepackage{makecell}
\usepackage{ragged2e}
 \usepackage{afterpage}
 \usepackage{capt-of}
 \definecolor{Gray}{gray}{0.9}
 \definecolor{LightCyan}{rgb}{0.88,1,1}
 
 \usepackage{booktabs}
\usepackage{multirow}
\usepackage{extarrows}
\usepackage{amssymb}

\newcommand{\indep}{\perp \!\!\! \perp}
\usepackage{epstopdf}
\usepackage{color}
\usepackage[dvipsnames]{xcolor}
\AtBeginEnvironment{quote}{\singlespacing\small}
\usepackage{tikz}

\newcommand{\task}{\textsc{ParCom}\xspace}
\newcommand{\method}{\textsc{SSP}lanner\xspace}

\newcommand*\circled[1]{\tikz[baseline=(char.base)]{
            \node[shape=circle,draw,inner sep=0.5pt] (char) {#1};}}

\newcommand*{\myalign}[2]{\multicolumn{1}{#1}{#2}}

\title{Plan ahead: Self-Supervised Text Planning \\for Paragraph Completion Task}

\author{Dongyeop Kang\thanks{$^*$This work was done while DK was at CMU.} \\ \texttt{dongyeopk@berkeley.edu} \\ \texttt{UC Berkeley}
        \And  
        Eduard Hovy \\ \texttt{hovy@cs.cmu.edu} \\ Carnegie Mellon University}

\begin{document}
\maketitle
\begin{abstract}
Despite the recent success of contextualized language models on various NLP tasks, language model itself cannot capture textual coherence of a long, multi-sentence document (e.g., a paragraph).
Humans often make structural decisions on what and how to say about before making utterances.
Guiding surface realization with such high-level decisions and structuring text in a coherent way is essentially called a planning process.
Where can the model learn such high-level coherence?
A paragraph itself contains various forms of inductive coherence signals called self-supervision in this work, such as sentence orders, topical keywords, rhetorical structures, and so on. 
Motivated by that, this work proposes a new paragraph completion task \task; predicting masked sentences in a paragraph.
However, the task suffers from predicting and selecting appropriate topical content with respect to the given context.
To address that, we propose a self-supervised text planner \method that predicts what to say first (content prediction), then guides the pretrained language model (surface realization) using the predicted content.
\method outperforms the baseline generation models on the paragraph completion task in both automatic and human evaluation. 
We also find that a combination of noun and verb types of keywords is the most effective for content selection. 
As more number of content keywords are provided, overall generation quality also increases. 
\end{abstract}

%---------------------------------%
\section{Introduction}\label{sec:intro}
%---------------------------------%

One may think textual coherence can be achieved from a gigantic language model trained on massive data.
This might be true in simple cases, such as generating short replies \cite{kannan2016smart}, but not in a long, multi-sentence generation.
This is mainly because per-word predictions from the autoregressive models can not capture the long-term flow of text, while humans often make structural decisions on what and how to say about before they speak \cite{byrne1979teaching,mckeown1985discourse,hovy1990pragmatics,gopen1990science,kang2020thesis}.
Guiding the surface-level realization with such high-level decisions and coherently structuring output text is called a \textit{planning} process.

Where can the model learn such high-level decisions related to long-term coherence?
A written paragraph itself can be a pot of golden resources, containing various forms of inductive coherence signals.
Different types of coherence signals in a paragraph have been studied and used in many different ways:
a sequence of words or sentences \cite{devlin2018bert,radford2019language}, a discourse structure of a text \cite{appelt1982planning,hovy1991approaches,kang2017detecting,kang19emnlp_flownet}, an order of sentences \cite{chambers2008unsupervised,barzilay2008modeling}, topic introduction, co-reference, a sequence of events \cite{tomkins1978script,schank2013scripts}, and more.
In this work, we primarily focus on the effect of topical content in text planning.

Despite the recent advances of contextualized language models \cite{devlin2018bert,radford2019language}, the lack of appropriate tasks makes it difficult to evaluate generation models' long-term coherence.
Prior tasks fall into classification or ranking problems, such as  narrative close task \cite{chambers2008unsupervised,mostafazadeh2016corpus}, sentence ordering \cite{barzilay2008modeling}, and next sentence prediction \cite{devlin2018bert}.
Some recent works focused on designing generation tasks: story generation \cite{fan-etal-2019-strategies}, text infilling \cite{huang2019inset,fedus2018maskgan,hua-wang-2019-sentence}, or paragraph bridging \cite{kang19emnlp_flownet}.
However, most of them suffer from predicting appropriate topical content given limited context, due to the limited usage of self-supervision signals from the paragraph. 

This work proposes a new open-ended paragraph completion task; \task; predicting the masked sentences in a paragraph.
Unlike the prior works, our task uses two effective ways of self-supervision learnt from a written paragraph itself:
(1) we augment more training instances via permutation masking and (2) resolve the context sparsity problem by providing a set of ground-truth content keywords and predicting them directly from context at testing time.

For the task, we propose a self-supervised text planner (\method) that explicitly predicts content keywords (content prediction) from context and guides the pretrained language model (surface-realization) using the predicted content.
The distribution of predicted keywords is then combined with the distribution of words in the language model using copy mechanism \cite{see2017get}. 
The predicted content keywords are an approximation of topical intents by the generator, providing a hint to guide the surface realizer to bridge the coherency gap between the given context and text to generate.
Overall, \method combines two advantages; micro-level language fluency from the pre-trained language model (bottom-up) and macro-level content choice controlled by the macro-level planning (top-down).
Our experiment shows that \method achieves significant improvements over the baselines in both automatic and human evaluation. 
%---------------------------------%
\section{Related Work}\label{sec:related}
%---------------------------------%

We first categorize a wide range of long-term coherent generation tasks (Table \ref{tab:comparison_content_systems}), based on their inclusion relationship (\textbf{C}-\textbf{T}) between the given context (\textbf{C}) and target to predict (\textbf{T}). 

\begin{table}[h]
\centering
\fontsize{7.8}{8.3}\selectfont{
\begin{tabular}{@{} c @{\hskip 0.05cm} c @{\hskip -0.2cm} c @{\hskip 0.05cm} c @{\hskip 0.05cm} c @{\hskip 0.05cm} c @{}}
\toprule
\textbf{C-T}     &   \textbf{Tasks}   &  \textbf{\makecell{Content\\Selection}} & \textbf{\makecell{Content\\Planning}}  & \textbf{\makecell{Content\\Ordering}} &  \textbf{\makecell{Surface\\Realization}}   \\
\midrule
$\subset$   & \myalign{@{}l}{Data-to-text} & \texttimes & \texttimes & \checkmark & \checkmark \\
 \hdashline[0.4pt/2pt]
$\supset$   & \myalign{@{}l}{Summarization} & \checkmark & \texttimes & $\triangle$ & \checkmark \\
\hdashline[0.4pt/2pt]
$\thickapprox$   & \myalign{@{}l}{Paraphrasing} & $\triangle$ & \texttimes & \texttimes & \checkmark \\
\hdashline[0.4pt/2pt]
$\indep$    & \makecell[l]{StoryGen, Text Infilling, \\Bridging, \task(ours)}& \checkmark & \checkmark & \checkmark & \checkmark \\
\bottomrule
\end{tabular}
}
\caption{\label{tab:comparison_content_systems} Comparison of generation tasks by different inclusion relationships between \textbf{C}ontext and \textbf{T}arget.}
\end{table}

\textbf{\texttt{C $\subset$ T}}:
Data-to-text produces text from structured data (e.g., table).
 \citet{moryossef-etal-2019-step,puduppully2019data,shen2019select,miculicich-etal-2019-selecting} combine content planning with surface realization. 
However, since content is explicitly provided as a data form, the planner mostly orders and structures, not prediction.

\textbf{\texttt{C} $\supset$ \texttt{T}}:
In abstractive summarization, all context information is entirely given in the source document, as a superset of target summaries to predict.
Thus, generation only pays attention to abstracting the context into a shorter form instead of content prediction or ordering.

\textbf{\texttt{C $\approx$ T}}:
Paraphrasing is transforming surface patterns of text while preserving its semantics.
\citet{fu2019paraphrase} used variational autoencoders for surface realization with a latent bag of words model for differentiable content planning, where content to generate itself is given in context, not requiring any content planning.

\textbf{\texttt{C $\indep$ T}}:
Story generation \cite{fan-etal-2019-strategies}, text infilling \cite{fedus2018maskgan,huang2019inset}, paragraph bridging \cite{kang19emnlp_flownet}, and our proposed \task are very challenging tasks where context and target have no overlap (open-ended), but they should be coherently connected.
Table \ref{tab:comparison_content_systems_2} categorize various generation models applied on the open-ended tasks (C$\indep$T), based on its self-supervision types:

\begin{table}[h]
\centering
\fontsize{8.0}{9.0}\selectfont{
\begin{tabular}{@{} c @{\hskip 0.02cm} c @{\hskip 0.02cm} c @{\hskip 0.05cm} c  @{\hskip 0.05cm} c @{}}
\toprule
 \makecell{\textbf{Models}\\($\indep$)}   &  \textbf{\makecell{Bidirect.\\Flow}} & \textbf{\makecell{Permutation\\ Masking}}  &   \textbf{\makecell{Content\\Guidance}}  &  \textbf{\makecell{Content\\Prediction}} \\
\midrule
  \myalign{@{}l}{\citet{keskarCTRL2019}} & \checkmark & \checkmark & \checkmark & \texttimes \\
  \hdashline[0.4pt/2pt]
  \myalign{@{}l}{\citet{fan-etal-2019-strategies}} & \texttimes & \texttimes &\checkmark  & \texttimes \\
  \hdashline[0.4pt/2pt]
  \myalign{@{}l}{\citet{huang2019inset}} & \checkmark & \checkmark & \texttimes & \texttimes\\
  \hdashline[0.4pt/2pt]
 \myalign{@{}l}{\citet{hua-wang-2019-sentence}} & \texttimes & \texttimes & \checkmark & \texttimes\\
 \hdashline[0.4pt/2pt]
\myalign{@{}l}{\citet{kang19emnlp_flownet}} & \checkmark & \texttimes & \texttimes & \texttimes \\
\hdashline[0.4pt/2pt]
\myalign{@{}l}{\method (ours)} & \checkmark & \checkmark & \checkmark & \checkmark \\
\bottomrule
\end{tabular}
}
\caption{\label{tab:comparison_content_systems_2} Comparison of generation models in \texttt{C}$\indep$\texttt{T} tasks by different self-supervision types.}
\end{table}

\citet{keskarCTRL2019} conditioned language models with topical words to control the target text.
\citet{fan-etal-2019-strategies} developed a surface realizer on anonymized entities using semantic role labeling.
\citet{hua-wang-2019-sentence} used pre-extracted topics to guide a generator to produce stylized argumentation text.
However, they are given the topical content as input (content guidance), while our \method directly predicts plan words from context (content prediction).

\citet{fedus2018maskgan,huang2019inset} developed various methods for text infilling task. 
Very similar to our task, \citet{kang19emnlp_flownet} developed language models informed by discourse relations on the bridging task; given the first and last sentences, predicting the intermediate sentences (bidirectional flow).
However, they did not explicitly predict content words given context nor use them as a self-supervision signal in training.
Unlike random masking in \citet{keskarCTRL2019,huang2019inset}, we propose a better data augmentation training method via permutation masking. 

%---------------------------------%
\section{\task: Paragraph Completion Task from Self-Supervision Signals}\label{sec:task}
%---------------------------------%

Our task is motivated by the recently proposed task; paragraph bridging \cite{kang19emnlp_flownet}, predicting intermediate sentences of a paragraph, given the first and the last sentences.
To prevent generation becoming too divergent from the context in story or prompt generation \cite{fan-etal-2019-strategies}, the bridging task restricts generation to end with the last sentence given, provided as an ending goal for generation. 

However, in the bridging task, the objective is to generate text by coherently linking the two extreme sentences, making the task itself too challenging even for human\footnote{METEOR score from human generation on the task is only about 4.5 \cite{kang19emnlp_flownet}}.
For instance, the first and last sentences are too sparse to generate multiple (from 2 to 5) target sentences, increasing divergence of generation exponentially.
Also, data usage in \cite{kang19emnlp_flownet} is very inefficient; training a single instance per paragraph.

To address those issues, we propose a new paragraph completion task \task by maximizing self-supervision presented in a paragraph itself (Figure \ref{fig:task_para_gen}).
We describe two types of self-supervisions: (1) masking a fixed-length of consecutive sentences in any position over a paragraph to maximize usage of a paragraph and (2) extracting partial keywords of the masked text as plan keywords to resolve the content sparsity problem.
Mainly, we learn the patterns between the context and the plan keywords in training and at testing time predict the plan keywords, and guide the surface generator (\S\ref{sec:method}).

\subsection{Data Augmentation via Permutation Masking}\label{sec:sentence_masking_permutation}
Our work is motivated by word masking, in training contextualized language models \cite{devlin2018bert}, but extending it to \textit{sentence-level} for learning longer coherence.

\begin{figure}[t!]
\vspace{0mm}
\centering
{
\subfloat[][
Sentence masking via permutation: t=1 (left) or t=2 (right):
One paragraph has a total of 5+4=9 training instances. 
]{
\includegraphics[trim=0cm 8.8cm 19.2cm 0cm,clip,width=.5\linewidth]{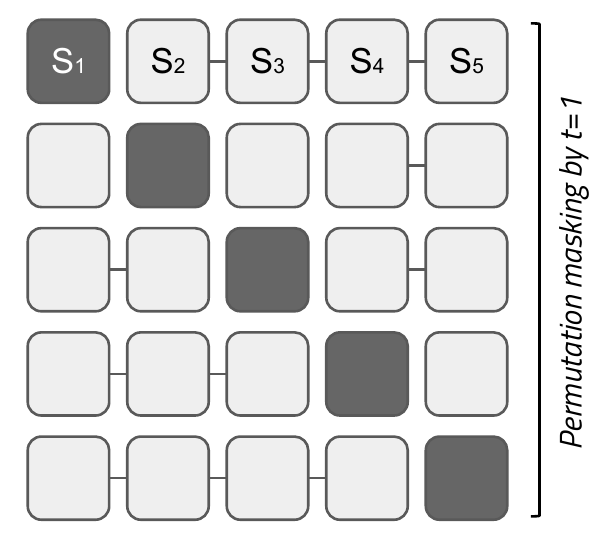}
\includegraphics[trim=7.0cm 8.8cm 12.2cm 0cm,clip,width=.5\linewidth]{{figs/textplan_task3}.pdf}
}
\\
\subfloat[\textwidth][Plan extraction on target sentence. The maximum number of keywords per sentence ($nkps$=2) is given.]{
\includegraphics[trim=0cm 4.7cm 19.8cm 6.1cm,clip,width=.65\linewidth]{{figs/textplan_task3}.pdf}
}
}
\caption{\label{fig:task_para_gen} Paragraph completion (\task) task: (a) predicting the \colorbox{black!60}{\color{white}{masked}}, target sentences given the \colorbox{gray!20}{unmasked}, context sentences  and (b) each masked, target sentence is given a small number of keywords extracted from the original target sentence. 
}
\end{figure}

Let $t$ be the number of targets, masked sentences to predict and $c$ be the number of unmasked, context sentences given, where $l$=$t$+$c$ is the total length of a paragraph.
For instance, in Figure \ref{fig:task_para_gen}, we have a $l$=5 length paragraph.
We restrict the number of context sentences to be larger than the number of target sentences ($c > t$), to avoid context become too sparse.
Also, we produce a total of 5+4=9 training instances, making use of data more efficient.

\subsection{Denser Context by Plan Extraction}\label{sec:plan_extraction}

We provide extra partial information as a set of keywords to guide the surface generator.
This is motivated by data-to-text tasks, but our plans are topical content instead of structured data.

We then question what types of plan keywords are the most effective for completing the paragraph.
We extract keywords using various keyword extraction systems: 
\begin{itemize}[noitemsep,topsep=0pt,leftmargin=*]
    \item \texttt{Off-the-shelf} systems extract keywords for each sentence using the three off-the-shelf systems:  YAKE \cite{campos2020yake} using statistical features (e.g., TF, IDF), RAKE \cite{rose2010automatic} using graph-based features (e.g., word degree), and PositionRank \cite{florescu2017positionrank} using position-based PageRank.
    Then we choose duplicate keywords by majority voting. 
    \item \texttt{Syntactic} features (e.g., part-of-speech tags, named entities \cite{fan-etal-2019-strategies}, events \cite{tomkins1978script}) are often regarded as the most salient topical content in generation.
    Using off-the-shelf Part-of-speech (PoS) tagger\footnote{\url{https://spacy.io/}}, we extract three types of syntactic features: \textit{nouns}, \textit{verbs}, and \textit{nouns+verbs}.
    \item \texttt{Attention} weights are used to capture context-aware keywords. We use the pre-trained BERT \cite{devlin2018bert} to encode context and target text, then average the attention weights of context words with respect to each target word. 
    We only use the first head's attentions, then average them over all 12 layers\footnote{ \citet{vig2019transformervis} observed that which layers or heads are important for syntactic and semantic tasks.}.
    We finally choose words with the maximum weight except for the special tokens (e.g., [CLS]) and punctuation marks.
\end{itemize}

We set the maximum number of keywords per sentence (\textit{nkps}) to 5. %to extract
Some extractors output an empty keyword list, so the number of keywords across the systems is different.
Our keywords are always uni-grams. 
In case they are not uni-grams, we split them by whitespaces and use individual unigrams as unique keywords. 
If the target text has multiple sentences, we combine all keywords from the sentences and randomly shuffle them.
The plan keywords extracted are only provided while training our plan predictor, but not at test time.
At testing time, we explicitly predict the keywords given context.
%---------------------------------%
\section{Self-supervised Text Planning (\method)}\label{sec:method}
%---------------------------------%
\begin{figure*}[th!]
\centering
{
\includegraphics[trim=0cm 6.5cm 9.8cm 0cm,clip,width=.96\linewidth]{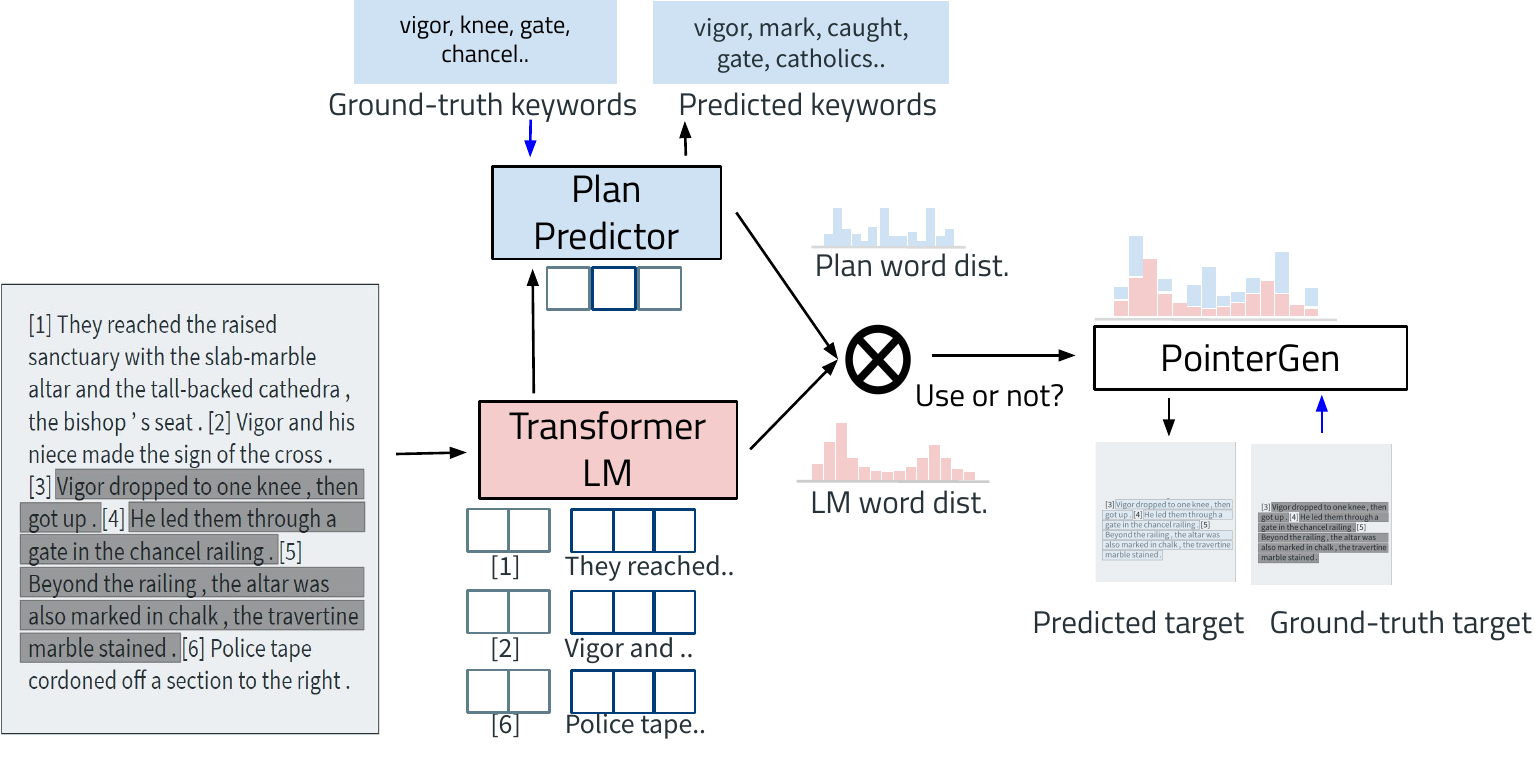}
}
\caption{\label{fig:planner} \method:  first predicts high-level plan keywords (Plan Predictor) then guides the surface generation (Transformer LM) using the predicted plan keywords.
The ground-truth plan keywords and target sentences (blue arrows) are only given in training time, whereas not in testing time.
The predicted and ground-truth target can be seen in Table \ref{tab:keywords_example}.
Best viewed in color.}
\end{figure*}

\method has various self-supervision modules that learn coherence signals from a paragraph itself: \textbf{surface realizer} (language model) by learning from a sequence of words, \textbf{next sentence predictor} by learning from a sequence of two consecutive sentences, \textbf{sentence position embeddings} by learning from an order of context sentences, \textbf{plan predictor} by learning from the relationship between the given context and important keywords used in the generation of the target text, and \textbf{content guidance} by learning from whether the predicted plan keywords are used or not in the target (See Figure \ref{fig:planner}).

Our planner is motivated by the two-stage generation framework \cite{moryossef-etal-2019-step,miculicich-etal-2019-selecting,fu2019paraphrase,hua-wang-2019-sentence}. While in prior works, the content is explicitly given from the dataset or task itself, our plan predictor in \method predicts the plan keywords only from the given context, by learning the topical relationship between context and content in target from training data.

Given $l$ length of a paragraph $s_1..s_l$ where each sentence $s$ consists of a $n$ number of words $s = w_{1}..w_{n}$, \task splits it into the context sentences $\texttt{x}$=$s_1..s_{j-1},s_{j+t}..s_n$ and $t$ target sentences to predict $\texttt{y}$=$s_j..s_{j+t-1}$.
For each target sentence, $p$ number of plan keywords $k_{j,1}..k_{j,p}$ for arbitrary target sentence $s_j$ are given only at training time.
The plan keywords are chosen from the entire vocabulary $\mathbb{V}^W$ and later combined with word distribution from  the language model.
We describe each self-supervision module in \method as follows:

%%%%%%%%%%%%%%%%%%%%%%%%%%%%%%%%%%%%
\textbf{Surface realization with pre-trained language models.}
%%%%%%%%%%%%%%%%%%%%%%%%%%%%%%%%%%%%
We use two different types of transformer-based language models: BERT \cite{devlin2018bert} and GPT2 \cite{radford2019language}.
While GPT2 is trained on bidirectionally tied language modeling, BERT is trained on masked language modeling. 
For BERT, we use the sequential sampling method \cite{wang2019bert}.
Using them, we encode context \texttt{x} and output the hidden representation $h_{j,i}  = \texttt{f} ( h_{j-1,i}, \texttt{x}_{k<(j,i)} )$ for $j^{th}$ word in $i^{th}$ sentence, where \texttt{f} $\in$ $\{$BERT, GPT2$\}$ is the transformer language model.
We then output the sentence vector $h_{i}$ by averaging all word vectors in a sentence.

%%%%%%%%%%%%%%%%%%%%%%%%%%%%%%%%%%%%
\textbf{Sentence position embedding.}
%%%%%%%%%%%%%%%%%%%%%%%%%%%%%%%%%%%%
We concatenate the encoded sentence representation with its sentence position embedding.
By adding the sentence position embeddings into context encoding, the model is aware of where the context sentence came from (e.g., from the first or last).
Compared to the simple concatenation of them \cite{kang19emnlp_flownet}, our sentence position embedding helps better learn the bi-directional coherence.
The context vector's final representation  is then $h^{c}  = \frac{1}{n}\sum_{i} h_{i} ; \texttt{pos}^{c}_i$ where $n$ is the number of sentences in a text and $\texttt{pos}^c_{i}$ is the position embedding of $i^{th}$ sentence in the context paragraph.

%%%%%%%%%%%%%%%%%%%%%%%%%%%%%%%%%%%%
\textbf{Plan prediction.}
%%%%%%%%%%%%%%%%%%%%%%%%%%%%%%%%%%%%
This work assumes that high-level plan words consist of bag-of-words \cite{fu2019paraphrase}, so that the model directly predicts the plan keywords from the vocabulary used in surface realization.
We calculate the plan probabilities over the entire vocabularies $\mathbb{V}$ given the context vector $h^{c}$ and choose the $p$ number of keywords with maximum probability estimates over vocabulary: $\hat{p}_{k\in \mathbb{V}}  = \texttt{softmax} (h^{c} W^{cv} \mathbb{V})$ where $\mathbb{V}$ is the vocabulary from the training data and $W^{cv}$ is the trainable model parameter. 
We do not control any explicit cut-off in the $p_{k\in \mathbb{V}}$ in order to make the distribution differentiable.
The objective is then:
\begin{align}
\mathbb{L}_{\texttt{plan}} &= - \sum_{k \in \mathbb{V}} \log p^{*}_k \log \hat{p}_k
\end{align}
where the loss is calculated by cross-entropy, $\hat{p}$ is the estimated probability distribution over vocabulary and $p*$ is the true one-hot distribution over plan keywords extracted from the extraction algorithms (i.e., [0,1..0,1] over $\mathbb{V}$).

%%%%%%%%%%%%%%%%%%%%%%%%%%%%%%%%%%%%
\textbf{Next sentence prediction.}
%%%%%%%%%%%%%%%%%%%%%%%%%%%%%%%%%%%%
Motivated by \citet{devlin2018bert}, we also add an auxiliary task of predicting whether the target sentence is related to context or not. 
For negative samples, \task assigns 50\% of random target sentences.
We optimize $\hat{p}_{next} = softmax( W^{c} h^c )$ where $W^{c}$ is the trainable parameter for the binary classification.
Next sentence prediction's objective is then:
\begin{align}
\mathbb{L}_{\texttt{next}} &= - \sum_{j} p^*_{next} \log \hat{p}_{next}
\end{align}
where the loss is calculated by binary cross-entropy, $p^*_{next}$ is the true label for next sentences and $\hat{p}_{next}$ is the predicted label.

%%%%%%%%%%%%%%%%%%%%%%%%%%%%%%%%%%%%
\textbf{Content guidance.}
%%%%%%%%%%%%%%%%%%%%%%%%%%%%%%%%%%%%
We combine two distributions between plan predictions and language modeling through copy mechanism following the pointer-generator \cite{see2017get}.
For $j^{th}$ sentence, we learn the probability of choosing the plan keyword or the word from language modeling based on context vectors, plan keyword distributions, and sentence position embedding of target sentences: $\texttt{P}_{plan} (v_k) = \sigma (  \mathsf{W}^{ck} [h^{c}; \hat{p_k}; \texttt{pos}^{t}_j] )$, where $\sigma$ is a sigmoid function, $\mathsf{W}^{ck}$ is the trainable parameter, and \texttt{v} $\in [0,1]$ is a probability of whether choosing the plan keyword or not.

We then decode each target sentence using the same language model decoder: $s_j = \texttt{g} (s_{j-1} , \hat{y}_{j-1})$, where $\texttt{g} \in \{$BERT, GPT2$\}$ is the language model decoder and $s$ is its output hidden state.
We can obtain the attention over plan keywords $k$:
\begin{align}
\alpha^{plan}_k &= \texttt{softmax}(\hat{p_k} \mathsf{W}^{kj} [s_j ; \texttt{pos}^t_j])
\end{align}
where $\mathsf{W}^{kj}$ is the trainable parameter.
Lastly, we combine the distribution of plan probabilities $\texttt{P}_{plan}$ and word probabilities in decoding $\texttt{P}_{lm}$.
%the distribution of 
\begin{align}
P(y) &= \texttt{P}_{plan}  \sum_k (\alpha^{plan}_k) + (1-\texttt{P}_{plan})  P_{lm} (y)
\end{align}
The objective of the pointer-generator is then:
\begin{align}
\mathbb{L}_{\texttt{gen}} & =  - \sum_{i\in\texttt{t} ,j=1..n} P(\hat{y}_{i,j}) \log P(y^{*}_{i,j})
\end{align}

%\taehee{SO MANY THEN!!}

%%%%%%%%%%%%%%%%%%%%%%%%%%%%%%%%%%%%
\textbf{Final objective.}
%%%%%%%%%%%%%%%%%%%%%%%%%%%%%%%%%%%%
The final objective of our training is to minimize the three objectives; plan prediction, next sentence prediction, and pointer generation, together:
\begin{align}
\mathbb{L}_{SPP} = \lambda_{plan} \mathbb{L}_{\texttt{plan}} +  \lambda_{next} \mathbb{L}_{\texttt{next}} +  \mathbb{L}_{\texttt{gen}}
\end{align}
where the weighting terms; $\lambda_{plan}$ and $\lambda_{next}$, are obtained through the cross-validation.

%---------------------------------%
\section{Experiment}\label{sec:exp}
%---------------------------------%

%hypotheses
We answer three questions in our experiments:
Q1. Does \method help produce a more coherent generation in \task? If so, which of the self-supervision modules are the most helpful? 
Q2. What types of plan keywords (e.g., noun, verb, attention) are most effective in terms of generation quality? How many keywords given are the most helpful?
Q3. Is \task a valid generation task to measure text coherence? 

%%%%%%%%%%%%%%%%%%%
\paragraph{Paragraph datasets.}
%%%%%%%%%%%%%%%%%%%

\begin{table}[h!]
\centering
\small
% \vspace{-2mm}
\begin{tabularx}{1.0\linewidth}{@{}l@{\hskip 0mm}c@{\hskip 1mm}c@{\hskip 1mm}c@{\hskip 1mm}c@{\hskip 1mm}c@{\hskip 1mm}c@{}}
\toprule
\textbf{Dataset} & \textbf{Domain} & \#\textbf{Sent.} & \#\textbf{Para.} & \textbf{Length}  & \#\textbf{Inst.} & \#\textbf{Keyw.}  \\
\midrule
% \texttt{SciFi}   & book & 825K       & 174K   & 4.7   & 10M      & 1.6M  & 1.6M/10M\\
\texttt{Fantasy} & book & 1.6M     & 352K   & 4.7   & 21M    &  3.2-19M\\
 \hdashline[0.4pt/2pt]
\texttt{Romance} & book & 5.3M     & 1.1M & 4.6   & 67M     & 27-62M\\
 \hdashline[0.4pt/2pt]
\texttt{WikiText} & wiki & 510K & 3.3M & 6.5 & 82M & 14M-78M \\
 \hdashline[0.4pt/2pt]
\texttt{CNNDM}   & news & 12M    & 311K   & 39.3  & 246M    &  63-315M\\
\bottomrule
\end{tabularx}
\caption{\label{tab:dataset} Data statistics: domain of text, the number of sentences, the number of paragraphs, the averaged length (number of sentences) of paragraph, the number of training instances permuted from the paragraphs, and minimum to maximum number of keywords extracted. 
}
\end{table}

\begin{table*}[h!]
\centering
\fontsize{9.0}{10.0}\selectfont{
\begin{tabularx}{0.99\linewidth}{@{}r c@{\hskip 2mm} c@{\hskip 2mm}c c@{\hskip 2mm}c@{\hskip 2mm}c c@{\hskip 2mm}c@{\hskip 2mm}c c@{\hskip 2mm}c@{\hskip 2mm}c@{}}
\toprule
   &   \multicolumn{3}{c}{\texttt{Fantasy}} & \multicolumn{3}{c}{\texttt{Romance}} & \multicolumn{3}{c}{\texttt{WikiText}} & \multicolumn{3}{c}{\texttt{CNNDM}} \\
\cmidrule(rr){2-4} \cmidrule(rr){5-7} \cmidrule(lr){8-10} \cmidrule(lr){11-13} 
\textbf{Models}     &  \textbf{B} & \textbf{M} & \textbf{VE}     &  \textbf{B} & \textbf{M} & \textbf{VE}     &  \textbf{B} & \textbf{M} & \textbf{VE}     &  \textbf{B} & \textbf{M} & \textbf{VE} \\
\midrule
\textbf{BiLSTM} \cite{hochreiter1997long}  & 2.6 & 2.9 & 26.5   & 2.2 & 2.4 & 25.6   & 2.7 & 2.9 & 31.2 &  2.5 & 2.5 & 30.0\\
\textbf{HRED} \cite{sordoni2015hierarchical} &  2.8 & 2.9 & 25.2   & 2.4 & 2.5 & 28.2   & 2.8 & 2.7 & 32.4 &  2.8 & 2.9 & 31.9\\
\textbf{FlowNet$_{disc}$} \cite{kang19emnlp_flownet} & 3.6 & 4.5 & 41.4  & 3.2 & 3.7 & 38.6   & 3.2 & 3.6 & 38.9   &  3.4 & 3.8 & 40.1 \\
\textbf{FlowNet$_{latent}$} \cite{kang19emnlp_flownet} &  3.7 & 4.5 & 42.8  & 3.1 & 3.6 & 35.2   & 3.1 & 3.5 & 37.5   & 3.3 & 3.7 & 38.7\\
\hdashline[0.4pt/2pt]
\textbf{BERT$_{finetune}$} \cite{devlin2018bert} & 3.7 & 4.6 & 38.5   & 4.1 & 4.4 & 42.8   &  4.2 & 4.7 & 48.9    &   4.4 & 4.8 & 47.5 \\
\textbf{GPT2$_{finetune}$} \cite{radford2019language} & 3.9 & 5.0 & 42.8   & 4.3 & 4.7 & 48.5 &  4.6 & 4.8 & 50.1   & 4.5 & 5.0  & 50.2\\
\hdashline[0.4pt/2pt]
\textbf{\method} (BERT) & 5.7 & 6.7 & 57.0   & 5.9 & 6.8 & 54.0   & 6.1  & 6.4 & 54.3    &  6.4 & 6.9 & 57.0\\
\textbf{\method} (GPT2) &     \textbf{7.1} & \textbf{9.2} & \textbf{69.5}   & \textbf{7.2} & \textbf{8.1} & \textbf{73.9}   &    \textbf{7.6} & \textbf{7.7} & \textbf{66.8}    & \textbf{6.9} & \textbf{7.8} & \textbf{59.9}\\
\hdashline[0.4pt/2pt]
\textbf{\method} (GPT2) $\backslash$w $\hat{p}$ &    11.1 & 11.8 & 79.6    &   12.7 & 13.3 & 84.4  &   12.5 & 13.0 & 87.8   & 12.1 & 12.9 & 84.9\\
\bottomrule
\end{tabularx}
}
\caption{\label{tab:result_generation_auto} Automatic evaluation.
\textbf{B} is BLEU, \textbf{M} is METEOR, and \textbf{VE} is vector extrema. 
For all metrics, the higher the better.
\method used keywords from the off-the-shelf system for training.
$\hat{p}$ is the ground-truth plan keywords extracted from the off-the-shelf system.
}
\end{table*}

Table \ref{tab:dataset} shows the paragraph datasets collected for our experiment.
We collect paragraphs from various domains: 
the two most frequent sub-genres extracted from BookCorpus \cite{moviebook} dataset; \texttt{Fantasy} and \texttt{SciFi}, Wikipedia text from \texttt{wikiText}-103 \cite{merity2016pointer}, and news articles from CNN/DailyMail (\texttt{CNNDM}) dataset \cite{see2017get}.
\texttt{CNNDM} and \texttt{WikiText} contain factual knowledge about events or things, whereas \texttt{Fantasy} and \texttt{Romance} are more narrative. 

For a fair comparison, we restrict the number of sentences in a paragraph from 4 to 7, the same as the setup in \citet{kang19emnlp_flownet}. %which is the
Since \texttt{CNNDM} has no specific line breakers in the document, each document is regarded as a single paragraph (39.3 lengths on average). 
Each dataset is randomly split by 0.9/0.05/0.05 for the train, valid, and test set, respectively.

%%%%%%%%%%%%%%%%%%%
\paragraph{Models.}
%%%%%%%%%%%%%%%%%%%
As baselines, we compare non-pretrained sequence-to-sequence models: \textbf{BiLSTM} \cite{hochreiter1997long} and hierarchical seq2seq \textbf{HRED} \cite{serban2017hierarchical,sordoni2015hierarchical} by encoding the concatenation of context sentences %regardless of their positions 
and then decoding the target sentences.
We also compare two strong paragraph generation models: \textbf{FlowNet$_{disc}$} using discourse relations and \textbf{FlowNet$_{latent}$} using latent delta relations \cite{kang19emnlp_flownet}, following the same setups (e.g., discourse parser, hyper-parameters) of the original paper.

Also, we use the pre-trained language model baselines fine-tuned on our paragraph datasets:
the fine-trained bert-base-uncased (\textbf{BERT}$_{finetune}$) and gpt2-base (\textbf{GPT2}$_{finetune}$) models \cite{Wolf2019HuggingFacesTS}.
For BERT, we use the sequential sampling method \cite{wang2019bert} with Nucleus sampling strategies for producing more diverse text \cite{Holtzman2019TheCC}.

Our proposed method \textbf{\method} is trained using either bert-base-uncased or gpt2-base.
As an upper-bound of our method, we predict masked, target text using the ground-truth plan keywords $\hat{p}$.

\paragraph{Setup.}
We find the best hyper-parameters on the validation set using a grid search on the learning rate, the number of training epochs, sampling parameters, and so on.
We follow the default parameters used in the HuggingFace's transformer models \cite{Wolf2019HuggingFacesTS}.
For a pointer-generator, we follow the default parameters in \cite{see2017get}.
The maximum number of plan keywords per sentence is 3. 

Here are the final parameters used for the \textbf{BiLSTM} and \textbf{HierLSTM} baselines:
$32$ for batch size, $128$ for maximum paragraph length, $300$ for word embedding size initialized by GloVe~\cite{pennington2014glove} for baseline models, $1$ LSTM layer \cite{hochreiter1997long} with $512$ size, clipping by $0.25$, $0.2$ learning rate and $0.5$ decay rate with Adagrad~\cite{duchi2011adaptive} optimizer, and $50,000$ for the vocabulary size.
For \textbf{FlowNet} variants, we follow the setup used in the original paper \cite{kang19emnlp_flownet}.
For BERT and GPT2 models, we use 32 for batch size, 2e-4 leraning rate with 1.0 maximum gradient norm and 0.02 weight decay using Adam~\cite{kingma2014adam} optimizer. 

Due to the computing limit, we restrict the maximum number of target sentences to 3 even though it could be up to half of the paragraph size in the full permutation.

\begin{table}[ht!]
\centering
\small
% \vspace{-2mm}
\begin{tabularx}{0.99\linewidth}{@{}r c@{\hskip 1mm}c@{\hskip 1mm}c c@{\hskip 1mm}c@{\hskip 1mm}c c@{\hskip 1mm}c@{\hskip 1mm}c@{\hskip 1mm}@{}}
\toprule
   &  \multicolumn{3}{c}{\texttt{Romance}} & \multicolumn{3}{c}{\texttt{WikiText}} & \multicolumn{3}{c}{\texttt{CNNDM}} \\
\cmidrule(rr){2-4} \cmidrule(rr){5-7} \cmidrule(lr){8-10} 
\textbf{Models}  &  \textbf{F} & \textbf{C} & \textbf{Q}     & \textbf{F} & \textbf{C} & \textbf{Q}     &  \textbf{F} & \textbf{C} & \textbf{Q}     \\
\midrule
\textbf{GPT2$_{finetune}$} & 4.4 & 2.1 & 3.6  &  3.9 & 1.9 & 1.9     &  3.8 & 1.6 & 1.8\\
\textbf{\method}  & 4.2 & 3.8 & 3.9     &  3.8 & 3.6 & 3.8   & 3.6 & 3.1 & 3.2\\
\textbf{\method} $\backslash$w $\hat{p}$ & 4.1 & 4.6 & 4.3   &  3.8 & 4.1 & 4.0 &   4.0 & 4.4 & 4.1\\
\midrule
\textbf{Human} & 4.8 & 4.9 & 4.9 &  4.6 & 4.5 & 4.5   & 4.5 & 4.4 & 4.4\\
\bottomrule
\end{tabularx}
\caption{\label{tab:result_generation_human} Human evaluation. \textbf{F} is fluency, \textbf{C} is coherence with context, and \textbf{Q} is overall quality. Each metric is scaled out of 5. 
}
\end{table}

%%%%%%%%%%%%%%%%%%%
\paragraph{Metrics.}
%%%%%%%%%%%%%%%%%%%
We evaluate our models using both automatic metrics and human evaluation:
For automatic metrics, we use two hard metrics: \textsc{BLEU} \cite{papineni2002bleu} and \textsc{METEOR} \cite{banerjee2005meteor}, as well as
an embedding similarity metric to capture the semantic similarity: Vector Extrema (\textsc{VE}) \cite{liu2016not}. 

For human evaluation, we measure fluency, coherence with respect to context, and overall quality with 1-5 Likert scale.
We randomly select 100 samples from the test set in each \texttt{Romance}, \texttt{WikiText}, and \texttt{CNNDM} (total 300 paragraphs).
Each sample is annotated by three crowd-workers then averaged.
We also measure how human performs on the task by asking workers to predict the masked text in these 300 paragraphs.

%%%%%%%%%%%%%%%%%%%%%%%%%%%%%%%%%%%%%%%%%%
\subsection{Automatic and Human Evaluation}
%%%%%%%%%%%%%%%%%%%%%%%%%%%%%%%%%%%%%%%%%%
Table \ref{tab:result_generation_auto} and \ref{tab:result_generation_human} show automatic and human evaluation result on \task task.
The fine-tuned models (\{BERT,GPT2\}$_{finetune}$)\footnote{In our experiment, no fine-tuned models (original pre-trained models) show very poor performance on our task.} and FlowNet models show significant improvements over the seq2seq baselines (BiLSTM and HRED) by large margins ($\sim$1.5 METEOR), showing the importance of fine-tuning on target text and modeling inter-sentential relation, respectively. 

In all datasets, \method shows significant improvements in both hard and soft metrics.
This indicates that explicitly predicting content words before surface realization helps generate more coherence text on target-oriented generation in \task.
\method with GPT2 outperforms \method with BERT, because such autoregressive models like GPT2 are more appropriate for our task, whereas BERT is not.
Finally, the performance of \method with the ground-truth keywords ($\hat{p}$) achieves the dramatic gain, which can be seen as an upper bound of our planning framework. 
Among domains, \texttt{Fantasy} and \texttt{Romance} seem to be better predicted compared to \texttt{WikiText} and \texttt{CNNDM} that require additional factual knowledge as well as narrative coherence.

Using the best model; \textbf{\method} (GPT2), we conduct a human evaluation on various system outputs and human-generated text (Table \ref{tab:result_generation_human}).
The fine-tuned GPT2 model shows high fluency as itself but very low coherence with context, because \task requires not only fluent and natural text but also context-aware text.
\method achieves much higher coherence and overall quality than the baselines, but still is far behind the upper-bound model (\method with $\hat{p}$) and human generation. %$\backslash$w

%%%%%%%%%%%%%%%%%%%%%%%%%%%%%%%%%%%%%%%%%%
\subsection{Performance of Self-supervision Modules}
%%%%%%%%%%%%%%%%%%%%%%%%%%%%%%%%%%%%%%%%%%
\begin{table}[t!]
\centering
\small
\begin{tabularx}{0.99\linewidth}{@{}l   cc cc cc@{}}
\toprule
   &   \multicolumn{2}{c}{\texttt{Romance}} & \multicolumn{2}{c}{\texttt{WikiText}} & \multicolumn{2}{c}{\texttt{CNNDM}} \\
\cmidrule(rr){2-3} \cmidrule(rr){4-5} \cmidrule(lr){6-7} 
\textbf{Models}    &  \textbf{NSP} & \textbf{PP}     &  \textbf{NSP} & \textbf{PP}     &   \textbf{NSP} & \textbf{PP}     \\
\midrule
\textbf{\method} &  91.6 & 48.1 & 92.7 & 50.2 & 90.7 & 49.4 \\
\bottomrule
\end{tabularx}
\caption{\label{tab:module_performance} Accuracies of each self-supervision module in \method.
\textbf{NSP} is next sentence prediction, and \textbf{PP} is plan prediction.
}
\end{table}

\begin{table*}[htbp!]
\centering
\small
\begin{tabularx}{1.0\linewidth}{@{}X@{}}
\textbf{Task}: given context sentences \circled{0}, \circled{1}, \circled{5}, predict target sentences \circled{2}, \circled{3}, \circled{4}\\
\toprule
\colorbox{gray!20}{\circled{0} ''They reached the raised sanctuary with the slab-marble altar and the tall-backed cathedra , the bishop ' s seat .'' \circled{1} ''Vigor} \colorbox{gray!20}{and his niece made the sign of the cross .''} \colorbox{black!60}{\color{white}{\circled{2} ''Vigor dropped to one knee , then got up .'' \circled{3} ''He led them through a gate in}} \colorbox{black!60}{\color{white}{ the chancel railing .'' \circled{4} ''Beyond the railing , the altar was also marked in chalk , the travertine marble stained .''}}\colorbox{gray!20}{\circled{5} ''Police } \colorbox{gray!20}{tape cordoned off a section to the right .''}\\
\\
\hdashline[0.4pt/2pt]
\\
\end{tabularx}
\begin{tabularx}{1.0\linewidth}{@{}rl@{}}
% \cmidrule(rr){3-6} \cmidrule(lr){7-8}
\multicolumn{2}{@{}l}{\textbf{Plan keywords} extracted from target sentences using different systems:}\\
\toprule
\texttt{Off-the-shelf} & \circled{2} (vigor, dropped, one), \circled{3} (chancel, railing, led), \circled{4} (travertine, marble, stained)\\
\texttt{Syntactic} (noun) & \circled{2} (vigor, knee), \circled{3} (gate, chancel), \circled{4} (railing, altar, chalk)\\
\texttt{Syntactic} (verb) &  \circled{2} (dropped, got),  \circled{3} (led, railing), \circled{4} (marked, stained)\\
\texttt{Syntactic} (nounverb) & \circled{2} (vigor, dropped, knee), \circled{3} (led, gate, chancel), \circled{4} (railing, altar, marked)\\
\texttt{Attention} &  \circled{2} (vigor, dropped, got), \circled{3} (led, gate, railing), \circled{4} (altar, chalk, travertine)\\
\\
\hdashline[0.4pt/2pt]
\\
\end{tabularx}
\begin{tabularx}{1.0\linewidth}{@{}p{1.8cm}p{6.7cm}p{6.7cm}@{}}
 &\textbf{\method} & \textbf{Human writer} \\
 \toprule
Human eval. & F : 4.3, C: 3.9, Q: 3.8 & F : 4.8, C: 4.9, Q: 4.8\\
\hdashline[0.4pt/2pt]
Predicted plan keywords & \circled{2} (vigor, mark, caught), \circled{3} (gate, catholics, police), \circled{4} (altar, mark, bishop)
& 
\circled{2} (vigor, show, sanctuary), \circled{3} (altar, blood, trace), \circled{4} (kill, sacrifice, recently) \\
\hdashline[0.4pt/2pt]
Predicted target sentences & 
\circled{2} ``vigor continuously walked down the road .'' \circled{3} ``he opened the gate which has a sign of catholics .'' \circled{4} ``both bishop and vigor met a police officer .''
& 
\circled{2} ``Then vigor showed around  the sanctuary to them.'' \circled{3} ``In there, they found a trace of the blood on the altar.'' \circled{4} ``They thought that  recently the sacrifice was killed in here.''\\
\\\hdashline[0.4pt/2pt]

\end{tabularx}
\caption{\label{tab:keywords_example} Example paragraph with the plan keywords extracted from different algorithms and output predictions by \method and human writer. \textbf{F} is fluency, \textbf{C} is coherence with context, and \textbf{Q} is overall quality. 
}
\end{table*}

We measure performance (i.e., accuracy) of each self-supervision module in \method: next sentence prediction (NSP) and plan prediction (PP) on the test samples (Table \ref{tab:module_performance}).
\method achieves very high accuracy in NSP.
In PP, \method correctly predicts almost half of the keywords from the total vocabulary size, indicating that the plan prediction module in \method can capture a certain level of coherence between the given context and target text to predict, although it is not perfect.

%%%%%%%%%%%%%%%%%%%%%%%%%%%%%%%%%%%%%%%%%%
\subsection{Comparison of Self-supervision Modules and Keyword Types in Training}
%%%%%%%%%%%%%%%%%%%%%%%%%%%%%%%%%%%%%%%%%%

\begin{table}[h!]
\centering\small
    \begin{tabular}{@{}l @{\hskip 3mm}  c@{\hskip 3mm}c @{}}
    \toprule
    \textbf{Models}    &  \textbf{M} & \textbf{VE}     \\
    \midrule
    \textbf{\method} & 7.9 & 66.6      \\
    - Sentence Position (SP) &  -1.4 & -8.1     \\
    - Plan Prediction (PP) &  -2.1 & -13.2     \\
    - Next Sentence Prediction (NSP) &  -0.5 & -3.6     \\
    \bottomrule
    \end{tabular}
\caption{\label{tab:ablation_modules} Ablation on self-supervision modules. 
}
\end{table}
Table \ref{tab:ablation_modules} shows ablation on self-supervision modules.
All scores are macro-averaged on three datasets: \texttt{Romance}, \texttt{WikiText}, and \texttt{CNNDM}.
Each module helps improve the overall performance (METEOR): plan prediction (+2.1 M), sentiment positional embedding (+1.4 M), and next sentence prediction (+0.4 M),

\begin{table}[h!]
\centering
\small
    \begin{tabular}{@{}l @{\hskip 3mm}  c@{\hskip 3mm}c @{}}
    \toprule
    \textbf{Models}    &  \textbf{M} & \textbf{VE}        \\
    \midrule
     $\backslash$w Random            & 6.1&	54.0\\
     $\backslash$w Syntac(Verb)      &7.6	&63.7 \\ 
     $\backslash$w Syntac(Noun)      & 7.5&	62.2 \\
     $\backslash$w Syntac(N+V) & \textbf{8.0} &	\textbf{66.3} \\
     $\backslash$w Off-the-shelf     & 7.8&	66.8\\
      $\backslash$w Attention         & 7.6&	63.6 \\
    \bottomrule
    \end{tabular}
\caption{\label{tab:ablation_plan_keyword_types} Comparison of plan keyword types at training (right). All scores are macro-averaged on three datasets: \texttt{Romance}, \texttt{WikiText}, and \texttt{CNNDM}.
}
\end{table}

Among the different types of keywords used in training (Table \ref{tab:ablation_plan_keyword_types}), the combination of nouns and verbs and the keywords extracted from the off-the-shelf algorithm outperform the other types.
We conjecture that since a sentence consists of both entities (i.e., nouns) and events (i.e., verbs) according to the script theory \cite{schank2013scripts}, the combination of them provides the largest amount of information to complete the sentence.
Attention-based keywords are not that helpful because the averaged attention weights themselves may not be a good indicator for topical coherence.

%%%%%%%%%%%%%%%%%%%%%%%%%%%%%%%%%%%%%%%%%%
\subsection{Comparison of Keyword Types and Ratios in Testing}
%%%%%%%%%%%%%%%%%%%%%%%%%%%%%%%%%%%%%%%%%%

\begin{figure}[h!]
\centering
% \vspace{-3mm}
\includegraphics[trim=0.3cm 0.6cm 1.4cm 0.2cm,clip,width=.53\linewidth]{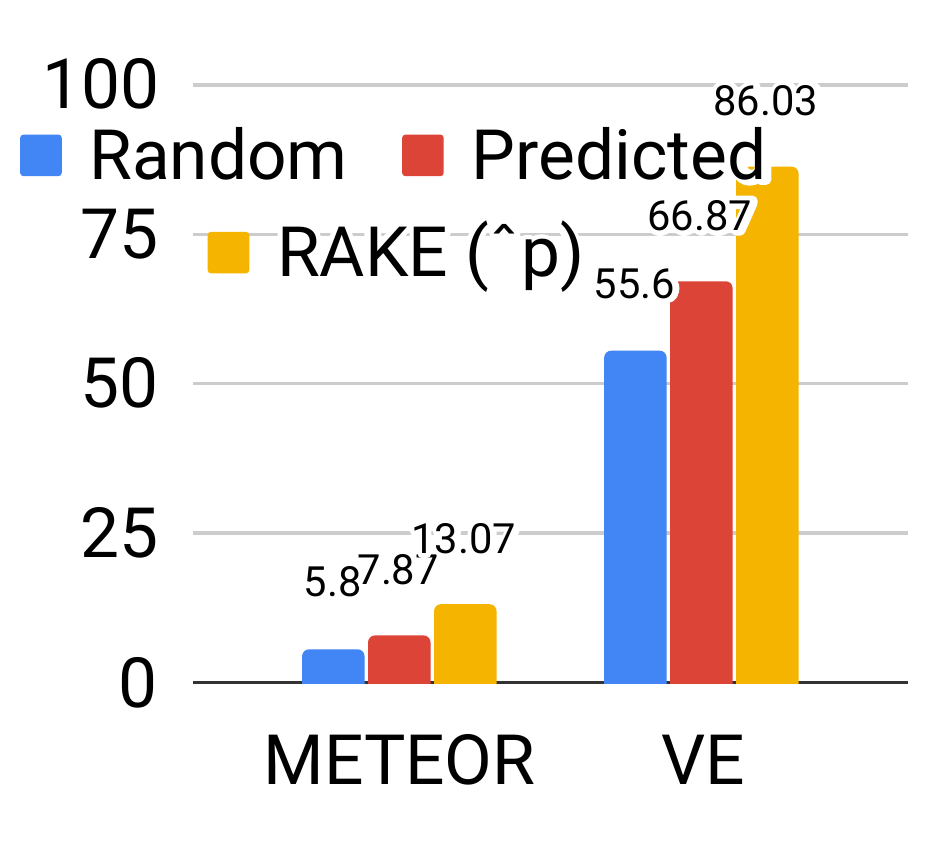}
\includegraphics[trim=1.8cm 0.6cm 1.0cm 0.2cm,clip,width=.455\linewidth]{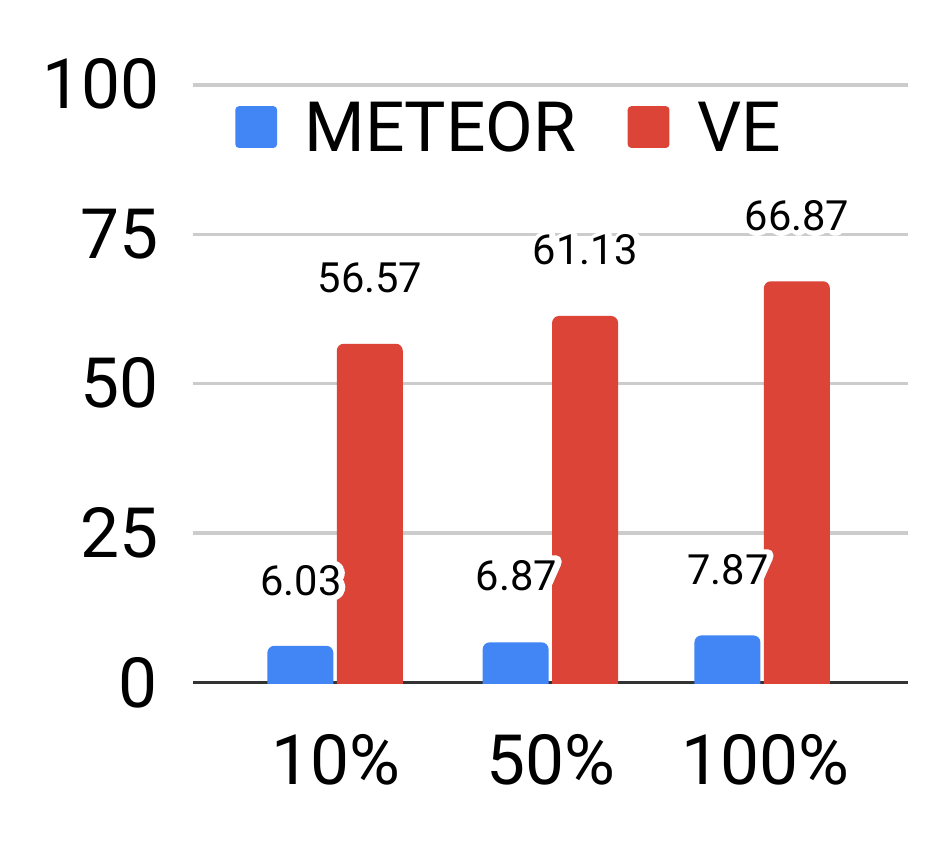}
\caption{\label{fig:ablation_keywords_testing} Comparison of plan keyword types (left) and plan keyword ratios used (right) in testing. Best viewed in color.
}
\end{figure}

In Figure \ref{fig:ablation_keywords_testing}, at test time, the predicted keywords from \method (red) shows dramatic improvements in both METEOR and VE against the random keywords (blue), but far behind the ground-truth keywords (yellow). 
As more predicted keywords are used at testing time, the generation quality increases.

Table \ref{tab:keywords_example} shows an example paragraph with ground-truth keywords extracted from different algorithms in \task and predicted target sentences with plan keywords by \method and a human writer.
In the prediction by \method, half of the predicted keywords are used in the generation, making the story more coherent to the first two sentences and the last ending sentence.
In the entire test set, we observe that about 43\% of predicted keywords are actually used in generation.

\section{Conclusion}\label{sec:conclusion}
A written paragraph itself contains various inductive coherence signals to be learned through \textit{self-supervision}.
Motivated by this, we propose a paragraph completion task for measuring textual coherence from a long document using different types of self-supervision signals.
To solve the task, we propose a text planner \method that explicitly predicts topical content keywords, and then guides the surface generator using the predicted plan keywords.
\method consists of different kinds of self-supervision modules: sentence positions, a sequence of words or sentences, and the topical relationship between context and target. 
Our self-supervised planning, in addition to other types of planning (e.g., discourse, goals, coreference, tenses) can be an important step toward modeling a long-term coherence in text generation.

Our results suggest several promising directions:
Although our ablation tests show the effect of each self-supervision module, types of plan keywords, and the amount of keywords with respect to generation quality, there are more spaces to explore in self-supervised text planning. For example, one can study the generation quality with respect to the position of the target sentences (beginning, middle, end), the comparison of plan keywords predicted by human and system, the effect of data augmentation by their positions (e.g., masking the only middle), the generation quality with respect to the ratio between masked and unmasked sentences, and more.

Second, we can extend the set of plan keywords to be more structured like a discourse tree. 
For instance, one can write a simple structure like ``(CAUSALITY (ELABORATE (Buy, Coffee)) (Pay, Tip, 12 dollars))'' then the system can generate a long, coherent text reflected by the structure. 
Predicting such structural plans from context and imposing them into the generator would be a potential direction for future work.

Last, text planning is a cognitive function commonly used in human language generation. 
To generate more human-like utterances, different planning stages should be simultaneously combined together \cite{kang2020thesis}, such as abstractive planning, strategic planning, coherence planning, and diversity planning. 
Combining the heterogeneous planning systems will be a crucial step towards developing a human-like language generation.

\section*{Acknowledgements} 
We thank Dery Lucio, Hiroaki Hayashi, Taehee Jung, Edvisees members at CMU, Hearst lab members at UC Berkeley, and anonymous reviewers at EMNLP 2020 for their helpful comments.

% \normalem
\bibliographystyle{acl_natbib}
\bibliography{main}

\begin{thebibliography}{46}
\expandafter\ifx\csname natexlab\endcsname\relax\def\natexlab#1{#1}\fi

\bibitem[{Appelt(1982)}]{appelt1982planning}
Douglas~E Appelt. 1982.
\newblock Planning natural-language utterances to satisfy multiple goals.
\newblock Technical report, SRI INTERNATIONAL MENLO PARK CA ARTIFICIAL
  INTELLIGENCE CENTER.

\bibitem[{Banerjee and Lavie(2005)}]{banerjee2005meteor}
Satanjeev Banerjee and Alon Lavie. 2005.
\newblock Meteor: An automatic metric for mt evaluation with improved
  correlation with human judgments.
\newblock In \emph{Proceedings of the acl workshop on intrinsic and extrinsic
  evaluation measures for machine translation and/or summarization}, pages
  65--72.

\bibitem[{Barzilay and Lapata(2008)}]{barzilay2008modeling}
Regina Barzilay and Mirella Lapata. 2008.
\newblock Modeling local coherence: An entity-based approach.
\newblock \emph{Computational Linguistics}, 34(1):1--34.

\bibitem[{Byrne(1979)}]{byrne1979teaching}
Donn Byrne. 1979.
\newblock \emph{Teaching writing skills}.
\newblock Longman.

\bibitem[{Campos et~al.(2020)Campos, Mangaravite, Pasquali, Jorge, Nunes, and
  Jatowt}]{campos2020yake}
Ricardo Campos, V{\'\i}tor Mangaravite, Arian Pasquali, Al{\'\i}pio Jorge,
  C{\'e}lia Nunes, and Adam Jatowt. 2020.
\newblock Yake! keyword extraction from single documents using multiple local
  features.
\newblock \emph{Information Sciences}, 509:257--289.

\bibitem[{Chambers and Jurafsky(2008)}]{chambers2008unsupervised}
Nathanael Chambers and Dan Jurafsky. 2008.
\newblock Unsupervised learning of narrative event chains.
\newblock In \emph{Proceedings of ACL-08: HLT}, pages 789--797.

\bibitem[{Devlin et~al.(2019)Devlin, Chang, Lee, and
  Toutanova}]{devlin2018bert}
Jacob Devlin, Ming-Wei Chang, Kenton Lee, and Kristina Toutanova. 2019.
\newblock Bert: Pre-training of deep bidirectional transformers for language
  understanding.
\newblock In \emph{NAACL}.

\bibitem[{Duchi et~al.(2011)Duchi, Hazan, and Singer}]{duchi2011adaptive}
John Duchi, Elad Hazan, and Yoram Singer. 2011.
\newblock Adaptive subgradient methods for online learning and stochastic
  optimization.
\newblock \emph{Journal of Machine Learning Research}, 12(Jul):2121--2159.

\bibitem[{Fan et~al.(2019)Fan, Lewis, and Dauphin}]{fan-etal-2019-strategies}
Angela Fan, Mike Lewis, and Yann Dauphin. 2019.
\newblock Strategies for structuring story generation.
\newblock In \emph{Proceedings of the 57th Annual Meeting of the Association
  for Computational Linguistics}, pages 2650--2660, Florence, Italy.
  Association for Computational Linguistics.

\bibitem[{Fedus et~al.(2018)Fedus, Goodfellow, and Dai}]{fedus2018maskgan}
William Fedus, Ian Goodfellow, and Andrew~M Dai. 2018.
\newblock Maskgan: better text generation via filling in the\_.
\newblock \emph{arXiv preprint arXiv:1801.07736}.

\bibitem[{Florescu and Caragea(2017)}]{florescu2017positionrank}
Corina Florescu and Cornelia Caragea. 2017.
\newblock Positionrank: An unsupervised approach to keyphrase extraction from
  scholarly documents.
\newblock In \emph{Proceedings of the 55th Annual Meeting of the Association
  for Computational Linguistics (Volume 1: Long Papers)}, pages 1105--1115.

\bibitem[{Fu et~al.(2019)Fu, Feng, and Cunningham}]{fu2019paraphrase}
Yao Fu, Yansong Feng, and John~P Cunningham. 2019.
\newblock Paraphrase generation with latent bag of words.
\newblock In \emph{Advances in Neural Information Processing Systems}, pages
  13623--13634.

\bibitem[{Hochreiter and Schmidhuber(1997)}]{hochreiter1997long}
Sepp Hochreiter and J{\"u}rgen Schmidhuber. 1997.
\newblock Long short-term memory.
\newblock \emph{Neural Computation}, 9:1735--1780.

\bibitem[{Holtzman et~al.(2019)Holtzman, Buys, Forbes, and
  Choi}]{Holtzman2019TheCC}
Ari Holtzman, Jan Buys, Maxwell Forbes, and Yejin Choi. 2019.
\newblock The curious case of neural text degeneration.
\newblock \emph{ArXiv}, abs/1904.09751.

\bibitem[{Hovy(1990)}]{hovy1990pragmatics}
Eduard~H Hovy. 1990.
\newblock Pragmatics and natural language generation.
\newblock \emph{Artificial Intelligence}, 43(2):153--197.

\bibitem[{Hovy(1991)}]{hovy1991approaches}
Eduard~H Hovy. 1991.
\newblock Approaches to the planning of coherent text.
\newblock In \emph{Natural language generation in artificial intelligence and
  computational linguistics}, pages 83--102. Springer.

\bibitem[{Hua and Wang(2019)}]{hua-wang-2019-sentence}
Xinyu Hua and Lu~Wang. 2019.
\newblock Sentence-level content planning and style specification for neural
  text generation.
\newblock In \emph{Proceedings of the 2019 Conference on Empirical Methods in
  Natural Language Processing and the 9th International Joint Conference on
  Natural Language Processing (EMNLP-IJCNLP)}, pages 591--602, Hong Kong,
  China. Association for Computational Linguistics.

\bibitem[{Huang et~al.(2019)Huang, Zhang, Elachqar, and Cheng}]{huang2019inset}
Yichen Huang, Yizhe Zhang, Oussama Elachqar, and Yu~Cheng. 2019.
\newblock Inset: Sentence infilling with inter-sentential generative
  pre-training.
\newblock \emph{arXiv preprint arXiv:1911.03892}.

\bibitem[{Kang(2020)}]{kang2020thesis}
Dongyeop Kang. 2020.
\newblock \emph{Linguistically Informed Language Generation: A Multifaceted
  Approach}.
\newblock {PhD} dissertation, Carnegie Mellon University.

\bibitem[{Kang et~al.(2017)Kang, Gangal, Lu, Chen, and
  Hovy}]{kang2017detecting}
Dongyeop Kang, Varun Gangal, Ang Lu, Zheng Chen, and Eduard Hovy. 2017.
\newblock Detecting and explaining causes from text for a time series event.
\newblock In \emph{Conference on Empirical Methods on Natural Language
  Processing}.

\bibitem[{Kang et~al.(2019)Kang, Hayashi, Black, and
  Hovy}]{kang19emnlp_flownet}
Dongyeop Kang, Hiroaki Hayashi, Alan~W Black, and Eduard Hovy. 2019.
\newblock Linguistic versus latent relations for modeling coherent flow in
  paragraphs.
\newblock In \emph{Conference on Empirical Methods in Natural Language
  Processing (EMNLP)}, Hong Kong.

\bibitem[{Kannan et~al.(2016)Kannan, Kurach, Ravi, Kaufmann, Tomkins, Miklos,
  Corrado, Luk{\'a}cs, Ganea, Young et~al.}]{kannan2016smart}
Anjuli Kannan, Karol Kurach, Sujith Ravi, Tobias Kaufmann, Andrew Tomkins,
  Balint Miklos, Greg Corrado, L{\'a}szl{\'o} Luk{\'a}cs, Marina Ganea, Peter
  Young, et~al. 2016.
\newblock Smart reply: Automated response suggestion for email.
\newblock In \emph{Proceedings of the ACM SIGKDD Conference on Knowledge
  Discovery and Data Mining (KDD)}, volume~36, pages 495--503.

\bibitem[{Keskar et~al.(2019)Keskar, McCann, Varshney, Xiong, and
  Socher}]{keskarCTRL2019}
Nitish~Shirish Keskar, Bryan McCann, Lav Varshney, Caiming Xiong, and Richard
  Socher. 2019.
\newblock {CTRL - A Conditional Transformer Language Model for Controllable
  Generation}.
\newblock \emph{arXiv preprint arXiv:1909.05858}.

\bibitem[{Kingma and Ba(2014)}]{kingma2014adam}
Diederik~P Kingma and Jimmy Ba. 2014.
\newblock Adam: A method for stochastic optimization.
\newblock \emph{arXiv preprint arXiv:1412.6980}.

\bibitem[{Liu et~al.(2016)Liu, Lowe, Serban, Noseworthy, Charlin, and
  Pineau}]{liu2016not}
Chia-Wei Liu, Ryan Lowe, Iulian~V Serban, Michael Noseworthy, Laurent Charlin,
  and Joelle Pineau. 2016.
\newblock How not to evaluate your dialogue system: {A}n empirical study of
  unsupervised evaluation metrics for dialogue response generation.
\newblock \emph{arXiv preprint arXiv:1603.08023}.

\bibitem[{McKeown(1985)}]{mckeown1985discourse}
Kathleen~R McKeown. 1985.
\newblock Discourse strategies for generating natural-language text.
\newblock \emph{Artificial Intelligence}, 27(1):1--41.

\bibitem[{Merity et~al.(2016)Merity, Xiong, Bradbury, and
  Socher}]{merity2016pointer}
Stephen Merity, Caiming Xiong, James Bradbury, and Richard Socher. 2016.
\newblock Pointer sentinel mixture models.
\newblock \emph{arXiv preprint arXiv:1609.07843}.

\bibitem[{Miculicich et~al.(2019)Miculicich, Marone, and
  Hassan}]{miculicich-etal-2019-selecting}
Lesly Miculicich, Marc Marone, and Hany Hassan. 2019.
\newblock Selecting, planning, and rewriting: A modular approach for
  data-to-document generation and translation.
\newblock In \emph{Proceedings of the 3rd Workshop on Neural Generation and
  Translation}, pages 289--296, Hong Kong. Association for Computational
  Linguistics.

\bibitem[{Moryossef et~al.(2019)Moryossef, Goldberg, and
  Dagan}]{moryossef-etal-2019-step}
Amit Moryossef, Yoav Goldberg, and Ido Dagan. 2019.
\newblock {S}tep-by-step: {S}eparating planning from realization in neural
  data-to-text generation.
\newblock In \emph{Proceedings of the 2019 Conference of the North {A}merican
  Chapter of the Association for Computational Linguistics: Human Language
  Technologies, Volume 1 (Long and Short Papers)}, pages 2267--2277,
  Minneapolis, Minnesota. Association for Computational Linguistics.

\bibitem[{Mostafazadeh et~al.(2016)Mostafazadeh, Chambers, He, Parikh, Batra,
  Vanderwende, Kohli, and Allen}]{mostafazadeh2016corpus}
Nasrin Mostafazadeh, Nathanael Chambers, Xiaodong He, Devi Parikh, Dhruv Batra,
  Lucy Vanderwende, Pushmeet Kohli, and James Allen. 2016.
\newblock A corpus and evaluation framework for deeper understanding of
  commonsense stories.
\newblock \emph{arXiv preprint arXiv:1604.01696}.

\bibitem[{Papineni et~al.(2002)Papineni, Roukos, Ward, and
  Zhu}]{papineni2002bleu}
Kishore Papineni, Salim Roukos, Todd Ward, and Wei-Jing Zhu. 2002.
\newblock Bleu: a method for automatic evaluation of machine translation.
\newblock In \emph{Proceedings of the 40th annual meeting on association for
  computational linguistics}, pages 311--318. Association for Computational
  Linguistics.

\bibitem[{Pennington et~al.(2014)Pennington, Socher, and
  Manning}]{pennington2014glove}
Jeffrey Pennington, Richard Socher, and Christopher Manning. 2014.
\newblock Glove: Global vectors for word representation.
\newblock In \emph{Proceedings of the 2014 conference on empirical methods in
  natural language processing (EMNLP)}, pages 1532--1543.

\bibitem[{Puduppully et~al.(2019)Puduppully, Dong, and
  Lapata}]{puduppully2019data}
Ratish Puduppully, Li~Dong, and Mirella Lapata. 2019.
\newblock Data-to-text generation with content selection and planning.
\newblock In \emph{Proceedings of the AAAI Conference on Artificial
  Intelligence}, volume~33, pages 6908--6915.

\bibitem[{Radford et~al.(2019)Radford, Wu, Child, Luan, Amodei, and
  Sutskever}]{radford2019language}
Alec Radford, Jeff Wu, Rewon Child, David Luan, Dario Amodei, and Ilya
  Sutskever. 2019.
\newblock Language models are unsupervised multitask learners.

\bibitem[{Rose et~al.(2010)Rose, Engel, Cramer, and Cowley}]{rose2010automatic}
Stuart Rose, Dave Engel, Nick Cramer, and Wendy Cowley. 2010.
\newblock Automatic keyword extraction from individual documents.
\newblock \emph{Text mining: applications and theory}, 1:1--20.

\bibitem[{Schank and Abelson(2013)}]{schank2013scripts}
Roger~C Schank and Robert~P Abelson. 2013.
\newblock \emph{Scripts, plans, goals, and understanding: An inquiry into human
  knowledge structures}.
\newblock Psychology Press.

\bibitem[{See et~al.(2017)See, Liu, and Manning}]{see2017get}
Abigail See, Peter~J Liu, and Christopher~D Manning. 2017.
\newblock Get to the point: Summarization with pointer-generator networks.
\newblock \emph{arXiv preprint arXiv:1704.04368}.

\bibitem[{Serban et~al.(2017)Serban, Sordoni, Lowe, Charlin, Pineau, Courville,
  and Bengio}]{serban2017hierarchical}
Iulian~Vlad Serban, Alessandro Sordoni, Ryan Lowe, Laurent Charlin, Joelle
  Pineau, Aaron~C Courville, and Yoshua Bengio. 2017.
\newblock A hierarchical latent variable encoder-decoder model for generating
  dialogues.
\newblock In \emph{AAAI}, pages 3295--3301.

\bibitem[{Shen et~al.(2019)Shen, Suzuki, Inui, Su, Klakow, and
  Sekine}]{shen2019select}
Xiaoyu Shen, Jun Suzuki, Kentaro Inui, Hui Su, Dietrich Klakow, and Satoshi
  Sekine. 2019.
\newblock Select and attend: Towards controllable content selection in text
  generation.
\newblock \emph{arXiv preprint arXiv:1909.04453}.

\bibitem[{Sordoni et~al.(2015)Sordoni, Bengio, Vahabi, Lioma, Grue~Simonsen,
  and Nie}]{sordoni2015hierarchical}
Alessandro Sordoni, Yoshua Bengio, Hossein Vahabi, Christina Lioma, Jakob
  Grue~Simonsen, and Jian-Yun Nie. 2015.
\newblock A hierarchical recurrent encoder-decoder for generative context-aware
  query suggestion.
\newblock In \emph{Proceedings of the 24th ACM International on Conference on
  Information and Knowledge Management}, pages 553--562. ACM.

\bibitem[{Swan(2002)}]{gopen1990science}
Judith~A. Swan. 2002.
\newblock The science of scientific writing.
\newblock In \emph{Book}.

\bibitem[{Tomkins(1978)}]{tomkins1978script}
Silvan~S Tomkins. 1978.
\newblock Script theory: Differential magnification of affects.
\newblock In \emph{Nebraska symposium on motivation}. University of Nebraska
  Press.

\bibitem[{Vig(2019)}]{vig2019transformervis}
Jesse Vig. 2019.
\newblock A multiscale visualization of attention in the transformer model.
\newblock \emph{arXiv preprint arXiv:1906.05714}.

\bibitem[{Wang and Cho(2019)}]{wang2019bert}
Alex Wang and Kyunghyun Cho. 2019.
\newblock Bert has a mouth, and it must speak: Bert as a markov random field
  language model.
\newblock \emph{arXiv preprint arXiv:1902.04094}.

\bibitem[{Wolf et~al.(2019)Wolf, Debut, Sanh, Chaumond, Delangue, Moi, Cistac,
  Rault, Louf, Funtowicz, and Brew}]{Wolf2019HuggingFacesTS}
Thomas Wolf, Lysandre Debut, Victor Sanh, Julien Chaumond, Clement Delangue,
  Anthony Moi, Pierric Cistac, Tim Rault, R'emi Louf, Morgan Funtowicz, and
  Jamie Brew. 2019.
\newblock Huggingface's transformers: State-of-the-art natural language
  processing.
\newblock \emph{ArXiv}, abs/1910.03771.

\bibitem[{Zhu et~al.(2015)Zhu, Kiros, Zemel, Salakhutdinov, Urtasun, Torralba,
  and Fidler}]{moviebook}
Yukun Zhu, Ryan Kiros, Richard~S. Zemel, Ruslan Salakhutdinov, Raquel Urtasun,
  Antonio Torralba, and Sanja Fidler. 2015.
\newblock Aligning books and movies: Towards story-like visual explanations by
  watching movies and reading books.
\newblock \emph{2015 IEEE International Conference on Computer Vision (ICCV)},
  pages 19--27.

\end{thebibliography}

\end{document}